\documentclass[runningheads]{llncs}

\usepackage{eccv}

\usepackage{eccvabbrv}

\usepackage{graphicx}
\usepackage{booktabs}
\usepackage{booktabs, tabularx} 
\usepackage{multirow}
\usepackage{multicol}
\usepackage{hyperref}
\usepackage{amsmath}  
\usepackage{algorithm}
\usepackage{algpseudocode} 
\usepackage[accsupp]{axessibility}  

\usepackage{hyperref}

\usepackage{orcidlink}

\begin{document}

\newcommand{\ccj}[2]{\textcolor{blue}{OLD: #1}\textcolor{orange}{NEW: #2} }

\title{VCD-Texture: Variance Alignment based 3D-2D Co-Denoising for Text-Guided Texturing}

\titlerunning{VCD-Texture}

\author{Shang Liu\inst{1,2} \and Chaohui Yu\inst{1,2} \and Chenjie Cao\inst{1,2} \and Wen Qian\inst{1,2} \and Fan Wang\inst{1,2}}


\institute{DAMO Academy, Alibaba Group \and
Hupan Lab \\
\email{\{liushang.ls, huakun.ych, caochenjie.ccj, qianwen.qian, fan.w\}@alibaba-inc.com}}

\maketitle

\begin{abstract}
Recent research on texture synthesis for 3D shapes benefits a lot from dramatically developed 2D text-to-image diffusion models, including inpainting-based and optimization-based approaches. However,  
these methods ignore the modal gap between the 2D diffusion model and 3D objects, which primarily render 3D objects into 2D images and texture each image separately.
In this paper, we revisit the texture synthesis and propose a \textbf{V}ariance alignment based 3D-2D \textbf{C}ollaborative \textbf{D}enoising framework, dubbed \textbf{VCD-Texture}, to address these issues.
Formally, we first unify both 2D and 3D latent feature learning in diffusion self-attention modules with re-projected 3D attention receptive fields. 
Subsequently, the denoised multi-view 2D latent features are aggregated into 3D space and then rasterized back to formulate more consistent 2D predictions.
However, the rasterization process suffers from an intractable variance bias, which is theoretically addressed by the proposed variance alignment, achieving high-fidelity texture synthesis.
Moreover, we present an inpainting refinement to further improve the details with conflicting regions.
Notably, there is not a publicly available benchmark to evaluate texture synthesis, which hinders its development. Thus we construct a new evaluation set built upon three open-source 3D datasets and propose to use four metrics to thoroughly validate the texturing performance.
Comprehensive experiments demonstrate that VCD-Texture achieves superior performance against other counterparts.

\end{abstract}

\begin{figure}
  \centering
  \includegraphics[width=1.0\textwidth]{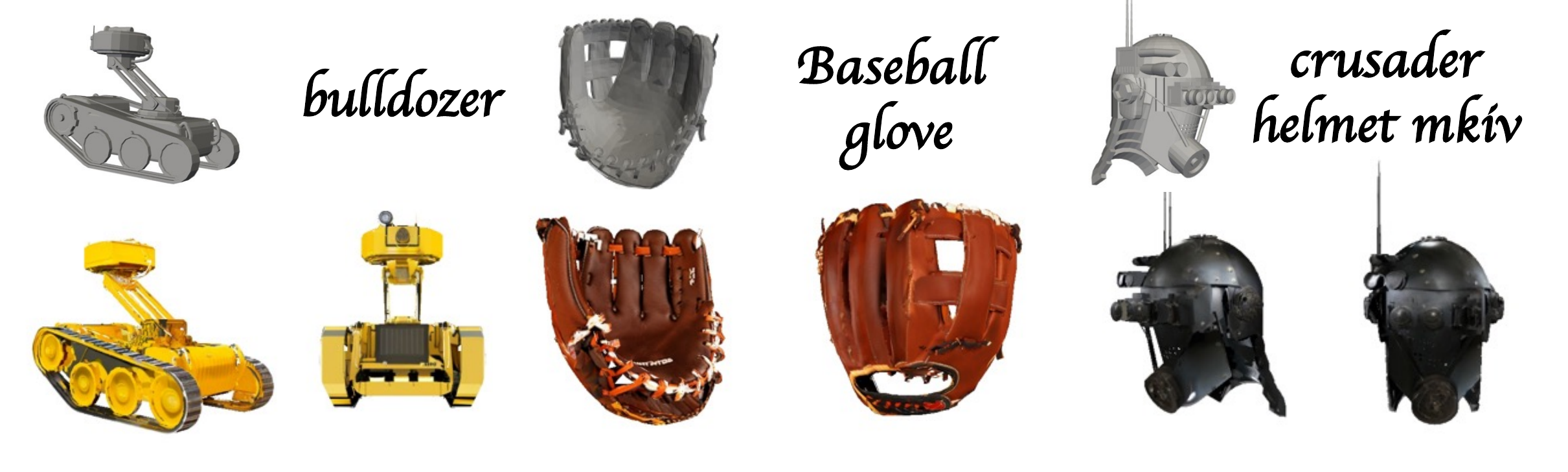}
  \vspace{-0.1in}
  \caption{Results of text-guided 3D shape textures generated by VCD-Texture. Our method could achieve high-quality texture synthesis with simple captions.
  }
  \label{fig:teaser}
  \vspace{-0.15in}
\end{figure}

\section{Introduction}
\label{sec:intro}

Textured 3D objects are essential in enhancing the realism and immersive experience in various computer graphics applications, including games, animation, and AR/VR environments~\cite{wei2004tile, games, kry2002eigenskin, kniss2005octree, chen20073d, cula20043d}. 
Traditionally, texture synthesis is a time-intensive and challenging task, often demanding specialized knowledge and extensive manual labor.
With prominent advancements in diffusion models~\cite{sd,saharia2022photorealistic,podell2023sdxl} trained with large-scale datasets~\cite{schuhmann2022laion, wang2022diffusiondb}, some recent researches~\cite{texfusion,Texture,chen2023text2tex,tango} have pivoted toward the adoption of 2D text-conditioned generative models, such as CLIP~\cite{clip} and Stable Diffusion (SD)~\cite{sd}, to direct the creation of 3D textures.
By leveraging the robust generalization properties inherent to pre-trained diffusion models, these methodologies enable to yield of high-quality textures.
For instance, 
prior optimization-based~\cite{clip-mesh, Text2Mesh} and inpainting-based approaches~\cite{repaint3D, Texture, chen2023text2tex} primarily render 3D objects into 2D images and texture each image through pre-trained 2D vision-language models. 
Very recently, driven by the advancement of 2D diffusion models, some literatures~\cite{texfusion,syncTexture,chen2023text2tex,repaint3D} further improve the texturing performance by designing texturing schemes in the image pixel domain or latent feature domain.

However, these text-to-image diffusion-enhanced texturing methods merely focus on synchronized multi-view diffusion denoising, neglecting cross-view correspondences in 3D space, as well as the fundamental disparity between the 2D diffusion model and 3D objects.
Moreover, we clarify that the feature aggregation-and-rasterization process~\cite{texfusion,syncTexture} suffers from a serious variance bias, which degrades the diffusion generation a lot.
Thus, in the shape texturing task, there remain challenges in seamlessly embedding 2D text-to-image diffusion priors to strengthen the 3D texture synthesis.

In this paper, we primarily address these issues through 1) \emph{unifying both 2D and 3D feature learning} during the text-to-image denoising process.
Furthermore, 2) we \emph{theoretically align the variance bias} that occurred in the aggregation-and-rasterization process for the multi-view feature fusion.
Finally, 3) we \emph{refine inconsistent regions with image inpainting}. 
Therefore, our method enjoys superior 3D consistency, unbiased high-frequency details, and high-fidelity textures as shown in Fig.~\ref{fig:teaser}.

Formally, we propose a novel 3D texture synthesis framework, called \textbf{V}ariance alignment based 3D-2D \textbf{C}ollaborative \textbf{D}enoising (\textbf{VCD-Texture}).
Particularly, {VCD-Texture} could be divided into two main processes: 3D-2D collaborative denoising and inpainting refinement. 
During the denoising steps of the Stable Diffusion (SD)~\cite{sd}, we modify the self-attention block in U-Net to learn features through both 2D and 3D respective fields with Joint Noise Prediction (JNP).
JNP utilizes the rendering-projection relationship to lift 2D foreground latent features into the 3D space, substantially improving the cross-view feature correlations.
Moreover, we propose to use Multi-View Aggregation-and-Rasterization (MV-AR) to fuse multi-view latent predictions from SD. 
Specifically, MV-AR is enhanced by a Variance Alignment (VA) technique to eliminate the domain gap of rasterization, which is utilized to re-render features from the aggregated 3D space to separate 2D views. 
To further improve the texture consistency, especially for the intrinsic discrepancy in latent feature and image pixel domains, we leverage the inpainting refinement to rectify inconsistent regions detected by aggregated pixels with high variance.

Benefiting from the integration of aforementioned components of {VCD-Texture}, our model could properly synthesize high-fidelity 3D textures.
However, the community still lacks a unified texture synthesis benchmark to evaluate the performance.
For quantitative experiments, we sample three sub-sets from publicly available 3D datasets (Objaverse~\cite{deitke2023objaverse}, ShapeNetSem~\cite{savva2015semgeo}, and ShapeNet~\cite{chang2015shapenet}), and leverage four metrics (FID~\cite{heusel2017gans}, ClipFID~\cite{kynkaanniemi2022role}, ClipScore~\cite{clip}, ClipVar~\cite{anonymous2023learning}) to thoroughly evaluate texture quality concerning fidelity, the semantic matching score between text and image, and the consistency of views.
Qualitative and quantitative comparisons signify that our {VCD-Texture} is capable of generating 3D textures with high fidelity, exhibiting both global and local coherence.
In addition, by harnessing the capabilities of pre-trained diffusion models, our approach, as a training-free method for 3D texture synthesis, demonstrates good time efficiency and remarkable and robust generalization ability in handling diverse 3D objects and complex textual descriptions.

The contributions of this paper are summarized as follows:
\begin{enumerate}
\item We propose a self-attention-based JNP block, which incorporates both 2D and 3D features to promote consistency of predicted multi-view noise.

\item We design MV-AR to generate consistent textures, and apply VA to theoretically address the rasterization variance bias problem.

\item We design a texture conflict identification and an inpainting refinement pipeline to mitigate the discrepancy between the feature domain and the pixel domain.
\end{enumerate}

\section{Related Works}
\label{sec:related_work}

\noindent\textbf{Text-to-Image Diffusion Models.}
In recent years, numerous advanced diffusion models~\cite{ramesh2022dalle2, sd, saharia2022photorealistic, zhang2023controlnet} have emerged, demonstrating their prowess in crafting high-fidelity images finely tuned by textual prompts. Among them, the immensely acclaimed SD~\cite{sd} stands out, which is honed on a diverse text-image compendium and intricately interwoven with the fixed text encoder from CLIP~\cite{clip}. The following works propose to add more conditions to text-to-image generation, including semantic segmentation~\cite{sd}, sketch~\cite{zhang2023controlnet}, depth map~\cite{zhang2023controlnet}, and other conditions~\cite{zhang2023controlnet,t2i-adapter}, which greatly promote the development and application of text-to-image generation.
Driven by the success of text-to-image diffusion models, many works have explored text-conditional diffusion models in other modalities, \emph{e.g.}, text-to-video~\cite{makeavideo}, and text-to-3D~\cite{dreamfusion}, and text-guided texturing~\cite{texfusion,chen2023text2tex,repaint3D}. In this work, we focus on the field of 3D texture generation.


\noindent\textbf{3D Shape and Texture Generation.}
Traditional Texture methods leverage rule-based or optimization-based approaches~\cite{ying2001texture, balas2006texture, chen2004texture, lefebvre2006appearance} to tile exemplar patterns to 3D assets. However, due to the limited computation and capacity, these methods are hard to synthesize complicated 3D models.
In the recent few years, deep learning methods~\cite{gan, stylegan} have been widely applied to the synthesis of 3D textures. 
Gramgan~\cite{portenier2020gramgan} trains deep models to generate textures by non-linearly combining learned noise frequencies.
SGAN~\cite{jetchev2016texture} projects a single noise vector to the whole spatial space via spatial GAN~\cite{gan} to synthesize texture.
Kniaz~\cite{kniaz2018thermal} proposes a novel method for the generation of realistic 3D models with thermal textures using the SfM~\cite{klodt2018supervising} pipeline and GAN.
Texture Fields~\cite{oechsle2019texture} utilizes a continuous 3D function parameterized with a neural network to achieve high-frequency textures. 
AUV-Net~\cite{chen2022auv} learns to map 3D texture into 2D plane by embedding 3D surfaces into a 2D aligned UV space.

Recently, with the development of the vision language model, recent texture research has explored techniques to distill language models for 3D texture generation.
Early approaches utilize CLIP~\cite{clip, x-mesh, Text2Mesh, clip-forge} to synthesize texture by improving the semantic similarity between rendering image and descriptive texture text.
Subsequently, given the advancement of text-to-image models, such as Stable Diffusion~\cite{sd}, DALLE~\cite{ramesh2022dalle2}, ControlNet~\cite{zhang2023controlnet}, which enables the generation of high-fidelity images with given description text.
To harness the capabilities of text-to-image models, TEXTure~\cite{Texture}, Text2Tex~\cite{chen2023text2tex} and Repaint3D~\cite{repaint3D} utilize Depth-aware Stable Diffusion (Depth-SD) to design an inpainting texturing scheme, which gradually paints the texture map of a 3D model from multiple viewpoints.
Later, Texfusion~\cite{texfusion} converts the progressive inpainting schema to the latent feature domain, which obtains consistent latent images via project-and-inpaint in each denoising time-step.
In further, SyncMVD~\cite{syncTexture} drops the auto-regressive inpainting schema and treats each view equally, which designs an aggregate-and-rendering process to synchronized multi-view latent features in each denoising time-step, and this allows the diffusion processes from different views to reach a consensus of the generated content early in the process and hence ensures the texture consistency.
Additionally, some studies, like Paint3D~\cite{paint3d} and UV-Diffusion~\cite{uv-diffusion}, involve training texture models from scratch. However, due to the limitations of 3D texture datasets, these models often exhibit suboptimal generalizability and produce textures that fall short of realism.

\section{Methods}
\label{sec:methods}
In this section, we first illustrate the preliminaries of diffusion models and mesh render principles in Sec.~\ref{sec:preliminary}. 
Then we detail about {VCD-Texture} overviewed in Fig.~\ref{fig:framework}(a).
In Sec.~\ref{sec:collaborative_denoising}, we introduce the 3D-2D collaborative denoising, including JNP with unified 3D-2D self-attention learning as depicted in Fig.~\ref{fig:framework}(b) and MV-AR enhanced with VA as shown in Fig.~\ref{fig:framework}(c) respectively.
Finally, the inpainting refinement detailed in Sec.~\ref{sec:inpainting_refinement} is utilized to further rectify inconsistent areas.

\subsection{Preliminary}
\label{sec:preliminary}
\subsubsection{Diffusion Models.}
The diffusion model~\cite{ddpm} comprises a forward process $q(\cdot)$ and a reverse denoising process $p_\theta(\cdot)$. The forward step incrementally corrupts data \( x_0 \) with noise \( \epsilon \sim \mathcal{N}(0, 1) \) to a noisy sequence \( x_1, \ldots, x_T \), following a Markov chain as:
\begin{equation}
    q(x_t | x_{t-1}, y) = \mathcal{N}(x_t; \sqrt{1 - \beta_t}x_{t-1}, \beta_t I),
     \quad
    x_t = \sqrt{\beta_t} x_0 + \sqrt{1 - \beta_t} \cdot \epsilon,
\end{equation}
where \( \beta_t \) is the variance schedule for \( t = 1, \ldots, T \); \( y \) is the condition.
The reverse process denoises the pure data from $x_T$ as:
\begin{equation}
    p_\theta(x_{t-1} | x_t, y) = \mathcal{N}(x_{t-1}; \mu_\theta(x_t, t, y), \Sigma_\theta(x_t, t, y)),
\end{equation}
where \( \mu_\theta \) and \( \Sigma_\theta \) indicate mean and variance predictions achieved by the trainable network parameterized with $\theta$.
\subsubsection{Stable Diffusion (SD).} 
\begin{figure}[tb]
  \centering
  \includegraphics[height=6.6cm]{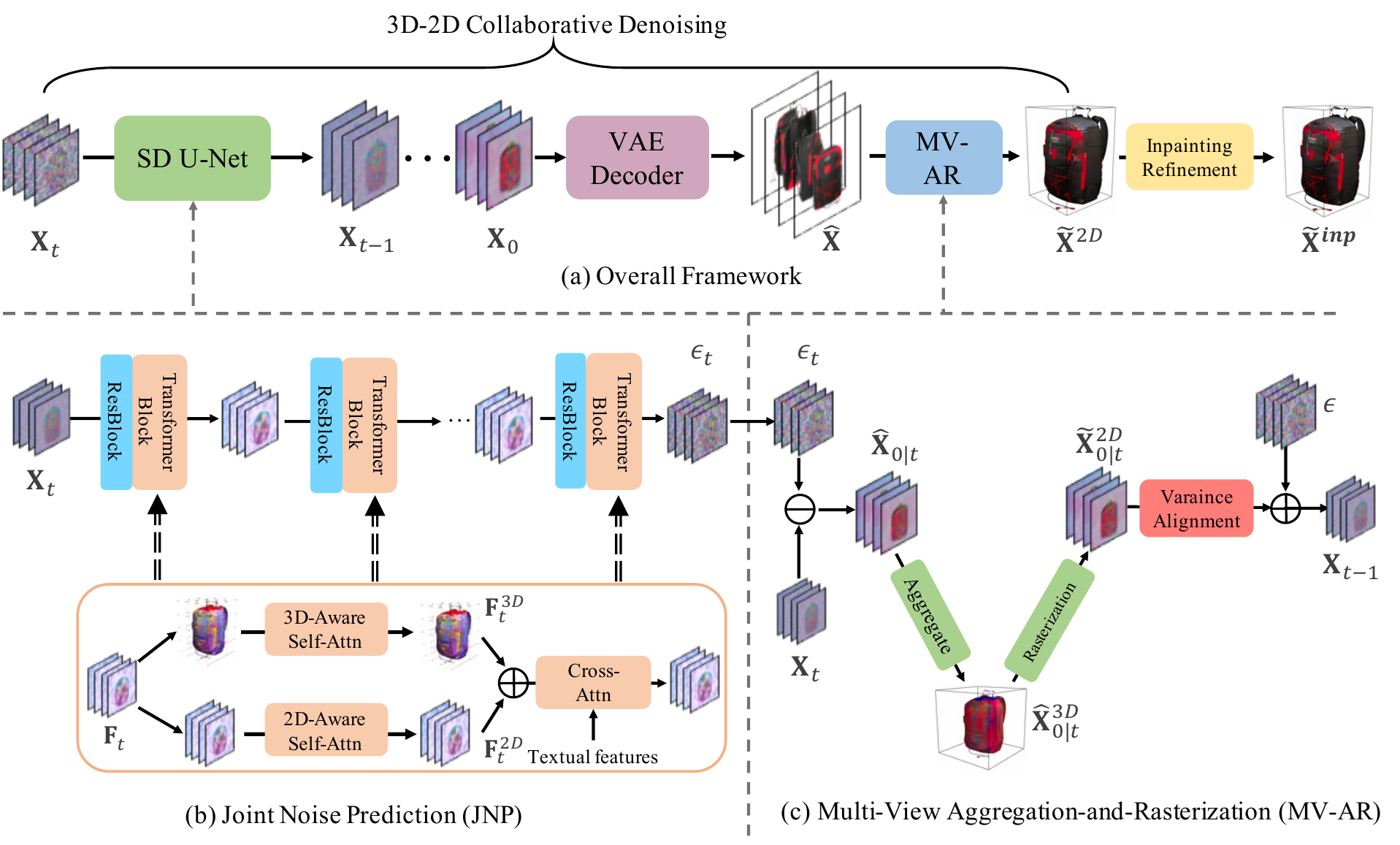}
  \vspace{-0.1in}
  \caption{The framework of VCD-Texture: (a) shows the overall process, including 3D-2D collaborative denoising and inpainting refinement; (b) shows the JNP in SD U-Net; (c) indicates the MV-AR with VA.
  Note that we only apply the aggregation sub-process of MV-AR to denoised multi-view images $\hat{\mathbf{I}}$ to achieve texture $\hat{\mathbf{I}}^{3D}$.
  }
  \label{fig:framework}
  \vspace{-0.15in}
\end{figure}

SD~\cite{sd}
projects all features into the latent space to save the diffusion computation. SD also employs an encoder-decoder-based U-Net for noise prediction. As shown in Fig.~\ref{fig:framework}(b), the multi-scale transformer block comprised in U-Net consists of self-attention and cross-attention modules. 
Self-attention lets the network prioritize long-range relevant features within a 2D image, while cross-attention aligns the denoising process with textual descriptions for controllable generation.

\subsubsection{Mesh Rendering~\cite{pineda1988parallel}.}
\label{subsubsec:Rendering}
Given a 3D mesh \( M \) with vertices \( V \), faces \( F \), and attributes \( A \), the rendering process converts the mesh into a 2D image by triangle rasterization. Each face \( f \in F \) is consisted of three vertices \( V_{f_1}, V_{f_2}, V_{f_3} \in V \) with attributes \( A_{f_1}, A_{f_2}, A_{f_3} \), where \( f_1, f_2, f_3 \) indicate indexes of vertices stored in face.
For each pixel value \( \mathbf{I}_{i} \), 
the rendering process entails tracing a ray that intersects the mesh at point \( P_{i} \) within triangle face \( f_k \).  Attribute of pixel \( \mathbf{I}_{i} \) are interpolated using barycentric coordinates \( (\alpha_{i1}, \alpha_{i2}, \alpha_{i3}) \), which satisfy:
\begin{equation}
\begin{aligned}
    \alpha_{i1} + \alpha_{i2} + \alpha_{i3} &= 1, \quad \alpha_{i1}, \alpha_{i2}, \alpha_{i3} \geq 0.
\end{aligned}
\label{eq:unit_constraint}
\end{equation}
Then, the pixel value \( \mathbf{I}_{i} \) is computed from a linear combination:
\begin{equation}
    \label{eq:rednering}
    \mathbf{I}_{i} = \alpha_{i1} A_{f1} + \alpha_{i2} A_{f2} + \alpha_{i3} A_{f3}.
\end{equation}
Given camera view \( C_{n\in N} \) with $N$ views at all, we aim to establish a mutual render-projection relation between 2D image plane \( \mathbf{I} \) and 3D mesh \( M \).
To achieve this, we utilize the above rendering approach to render \( M \) at the same size as image \( \mathbf{I} \). 
Thereby, we obtain rendering mapping function \( R(\cdot)\), which rasterizes mesh vertices to the 2D image plane, and back projection function \( R^{-1}(\cdot)\), which projects 2D plane values back to corresponding vertices.

\subsection{3D-2D Collaborative Denoising}
\label{sec:collaborative_denoising}

Given mesh $M$, text condition $y$, and camera $C_{n\in N}$ with $N$ views, we first re-mesh the initial mesh into a coarse one as $M_c$~\cite{repaint3D}, containing $J_c$ vertices. 
Next, we render $M_c$ into a latent space, which allows us to align the VCD-Texture learning process to the formulation of the SD U-Net.

\subsubsection{Joint Noise Prediction (JNP).}
\label{sec:JNP}
During the denoising step of time step $t$ in U-Net, we indicate $N$ views' input feature tensor to each transformer block as $\mathbf{F}_{t}\in \mathbb{R}^{N\times\hat{h}\times\hat{w}\times{c}}$, where $\hat{h},\hat{w},c$ are height, width, and channels of the feature.
Specifically, {VCD-Texture} unifies both 2D and 3D self-attention learning for diffusion denoising to achieve consistent feature presentation indicated as $\mathbf{F}_t^{2D}$ and $\mathbf{F}_t^{3D}$ respectively. 
Since $\mathbf{F}_t^{2D}$ can be separately learned by the inherent 2D self-attention of SD for each view, we incorporate another 3D self-attention branch as shown in Fig.~\ref{fig:framework}(b) with tailored 3D receptive fields for $\mathbf{F}_t^{3D}$.
Formally, we first apply the inverse projection \( R^{-1}(\cdot)\) to obtain the spatial mapping of every foreground feature to the 3D space. Then we split the 3D space into a series of discrete volumetric grids with grid size $G_t$. In the 3D attention branch, multi-view features $\mathbf{F}_t^{3D}$ can only attend to the ones within the same 3D grid, but it enjoys cross-view information which is ignored in $\mathbf{F}_t^{2D}$. 
Note that we empirically found that the whole process is training-free and the tensor shape can be also retained because we only adjust the attention receptive fields of $\mathbf{F}_t^{3D}$ while all parameters are frozen. More implementation details about the 3D self-attention are discussed in the supplementary.
After that, we average $\mathbf{F}_t^{2D}$ and $\mathbf{F}_t^{3D}$ as the final output of self-attention, which considers both globally long-range consistency for each separate 2D view and locally cross-view correlations.

Since the non-overlapping grid-based splitting with the same grid size suffers from limited feature interaction~\cite{liu2021swin}, we interactively two different grid sizes $G_t$ across the whole denoising process with different time steps. Thus, our model could eliminate the isolation with overlapped 3D receptive fields.
\subsubsection{Multi-View Aggregation-and-Rasterization (MV-AR).}
\label{sec:MV-AR}

To fuse the multi-view latent predictions $\mathbf{X}_{n\in N}$, we first compute the view score $S_{n\in N}$, which quantifies the cosine similarity between the mesh normal and the screen view direction. Note that we omit the time step $t$ for simplicity. 
The distance score \( D_{n\in N} \) is computed by: \( 1 - d_i / Z_{far} \), where \( d_i \) denotes distance between pixel $i$ and mesh surface,  \( Z_{far}\) represent the upper bound of the whole scene.
Then we derive the barycentric coordinate map \( B_{n\in N} \in \mathbb{R}^{h \times w \times 3} \) from \( R(\cdot)\), where channels are denoted by \( (\alpha_{i1}, \alpha_{i2}, \alpha_{i3}) \) respectively.
Subsequently, we formulate the vertex latent features $\hat{\mathbf{X}}_{n,j}\in\mathbb{R}^{4}$ and vertex weights $W_{n,j}\in\mathbb{R}^{1}$ for view $n\in N$, vertex index $j\in J_c$ as:
\begin{align}
    \hat{\mathbf{X}}_{n,j} &= \sum^{hw}_{i=1} \mathbf{X}_{n,i}  \cdot \psi(B_{n,i}, \tau_b) / \eta,\\
    W_{n,j} &= \sum^{hw}_{i=1} S_{n,i} \cdot \psi(D_{n,i}, \tau_d) \cdot \psi(B_{n,i}, \tau_b) / \eta,
    \label{eq:2d_to_vertice}
\end{align}
where $i\in hw$ indicate the 2D pixel indices; \(\eta=\sum^{hw}_{i=1} \psi(B_{n,i}, \tau_b)\); \( \psi \) represents power function; both \( \tau_b \) and \( \tau_d \) denote the exponent. 
Furthermore, we aggregate $\hat{\mathbf{X}}_{n,j}$ into $\hat{\mathbf{X}}^{3D}_{j}$ across all $N$ views as:

\begin{equation}
    \hat{\mathbf{X}}^{3D}_{j} = \sum^{N}_{n=1} \hat{\mathbf{X}}_{n,j}  \cdot \psi(W_{n,j}, \tau_w) / \omega, \quad \omega=  \sum^{N}_{n=1} \psi(W_{n,j}, \tau_w).
    \label{eq:2d_to_vertice_agg}
\end{equation}
With the aggregated 3D vertex feature $\hat{\mathbf{X}}^{3D}_j$, we further rasterize them back to the 2D plane with initial camera views and replace the foreground rendering area in predicted latent features $\mathbf{X}_{n}$ with the re-rendered ones, resulting in $\tilde{\mathbf{X}}_{n}^{2D}$. 
As 2D latent features $\tilde{\mathbf{X}}_{n}^{2D}$ are rendered from the 3D feature mesh, they naturally achieve superior view consistency.
We subsequently add the noise to the features to produce the latent features for step $t-1$, as the diffusion iteration~\cite{sd} displayed in Fig.~\ref{fig:framework}(c).

\subsubsection{Variance Alignment (VA).}
\label{sec:VA}

In diffusion models, controlling the proper stepwise variance through a designated noise schedule is critical for maintaining stability and producing high-quality images.
However, we find that the rasterization process in latent space would cause variance degradation during the denoising, resulting in over-smoothed generations.

To provide an in-depth analysis of this phenomenon, we begin by examining the rendering formula (Eq.~\ref{eq:rednering}) and the unit constraint condition (Eq.~\ref{eq:unit_constraint}). We observe that the rasterization process can be represented as a convex combination, which follows the same linear combination formula and coefficient conditions.
Furthermore, the variance $Var(\cdot)$ represents the expectation of a square function, which is naturally convex. Thus the variance satisfies the fundamental formula and conditions of Jensen's inequality as the convex function and the convex combination.
Formally, Jensen's inequality states that if $\varphi(\cdot)$ is a convex function, and \( z_{i\in N_z} \) are points in interval \( Z \), where \(N_z\) is the number of sampling points, then for any non-negative weights \( \lambda_i\) that satisfy the condition: \( \sum^{N_z}_{i=1} \lambda_i = 1 \), the following inequality holds:
\begin{align}
    \varphi\left(\sum_{i=1}^{N_z} \lambda_i z_i\right) \leq \sum_{i=1}^{N_z} \lambda_i \varphi(z_i).
    \label{eq:jensen_inequality}
\end{align}
Thereby, for random variable set $\mathbf{x}_i$, and any non-negative weights \(\lambda_i\) satisfying the condition \( \sum^{N_z}_{i=1} \lambda_i = 1 \), we can indicate:
\begin{align}
    Var\left(\sum_{i=1}^{N_z} \lambda_i \mathbf{x}_i\right) \leq \sum_{i=1}^{N_z} \lambda_i Var(\mathbf{x}_i).
    \label{eq:inequality}
\end{align}
This means that the variance of the convex value combination is no larger than the convex combination of variance, and this inequality is deduced in the supplementary materials in detail.

Revisiting the rasterization process, 
we claim that the rasterized 2D latent feature $\tilde{\mathbf{X}}$ contain smaller variance compared to the aggregated one $\hat{\mathbf{X}}^{3D}$ as:
\begin{align}
    Var(\tilde{\mathbf{X}}^{2D}) \leq \sum_{u=1}^{3} Var(\hat{\mathbf{X}}^{3D}_u) \cdot B^{3D}_u = Var(\hat{\mathbf{X}}^{3D}),
    \label{eq:inequality_3d2d}
\end{align}
where $B^{3D}_u$ indicates three barycentric coordinate banks with $u\in[1,2,3]$ triangle indices of each face. 
To address this issue, 
we propose a variance correction formulation defined as follows:
\begin{align}
\tilde{\mathbf{X}}^{2D'} = \frac{\tilde{\mathbf{X}}^{2D} - \mu(\tilde{\mathbf{X}}^{2D})}{\delta^{2D}} \cdot {\delta^{3D}} + \mu(\hat{\mathbf{X}}^{3D}),
\label{eq:std_correct}
\end{align}
where \(\mu(\cdot)\) represents the mean operation; \(\delta\) denotes standard deviation. And \( \delta^{3D} = \sqrt{Var(\tilde{\mathbf{X}}^{3D})} \) is achieved by:
\begin{align}
Var(\tilde{\mathbf{X}}^{3D}) &= \sum_{u_1=1}^{3} ({B^{3D}_{u_1}})^2 \cdot Var(\hat{\mathbf{X}}^{3D})\\
&+\sum^{3}_{u_1,u_2=1\atop u_1 \neq u_2} 
2 \cdot B^{3D}_{u_1} \cdot B^{3D}_{u_2} * Cov(\hat{\mathbf{X}}^{3D}_{u_1}, \hat{\mathbf{X}}^{3D}_{u_2}),
\label{eq:variance}
\end{align}
where \( Cov(\cdot, \cdot) \) represents covariance function. 
Eq.~\ref{eq:std_correct} meticulously aligns the feature variance after the rasterization to the mesh feature $\hat{\mathbf{X}}^{3D}$, which is capable of synthesizing realistic textures with more high-frequency details. We also empirically verified the effectiveness of VA in Fig.~\ref{fig:variance} and related ablation studies of Sec.~\ref{sec:abl}.

\subsection{Inpainting Refinement}
\label{sec:inpainting_refinement}
Due to the intrinsic discrepancy in resolution size and channel number between the latent and image domains, consistent latent cannot guarantee consistent images.
In this section, we introduce inpainting refinement to further alleviate the inconsistency in pixel values across different views of the image domain. 
Similar to the latent aggregation process, we generate the initial texture via pixel domain aggregation.
\begin{figure}[tb]
  \centering
  \includegraphics[width=9.0cm]{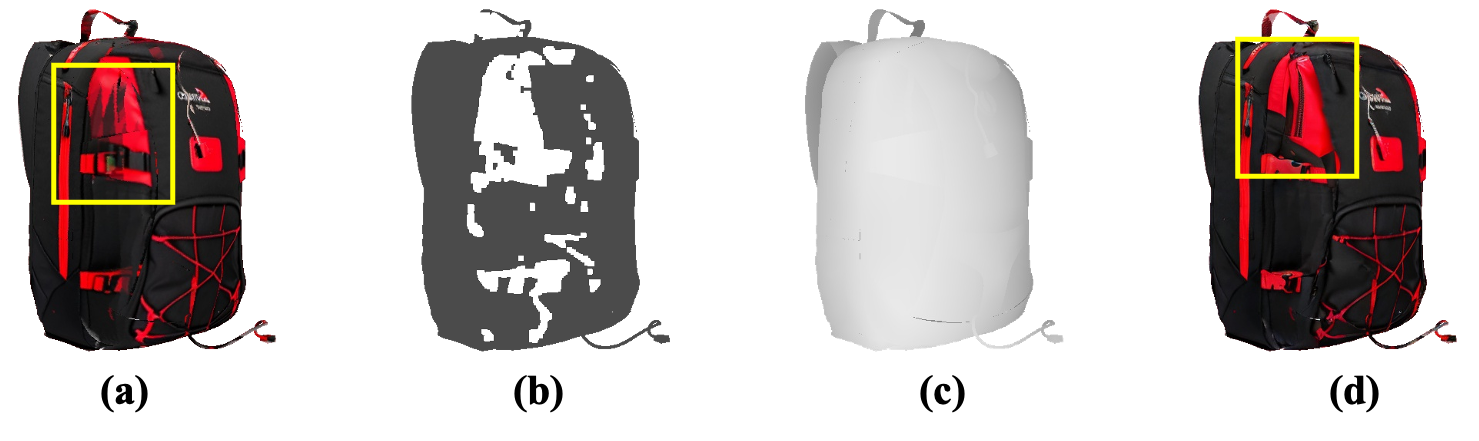}
  \vspace{-0.1in}
  \caption{The illustration of inpainting refinement. (a) shows the image view rendered from the initially inconsistent texture $\hat{\mathbf{I}}^{3D}$; (b) represents the dilated inpainting mask rendered from 3D mask $\mathbf{M}$; (c) is the depth map rendered from the input mesh; (d) indicates the updated final texture through our inpainting refinement. }
  \label{fig:inpainting}
  \vspace{-0.15in}
\end{figure}
We first identify those inconsistent vertices by the variance of related aggregation pixels. Then, we incorporate pixel domain texture inpainting approach~\cite{repaint3D} to refine those inconsistent regions, and all intermediate results of the inpainting refinement are shown in Fig.~\ref{fig:inpainting}. 
Specifically, given predicted multi-view images $\mathbf{I}_{n_f\in N_f}$, where $N_f=4$ views are sampled from the total $N$ views, we calculate the variance of vertex $j$ as:
\begin{align}
Var_j = \sum^{N_f}_{n_f=1} (\hat{\mathbf{I}}^{3D}_{j,n_f} -\mu(\hat{\mathbf{I}}^{3D}_{j}))^2 / (N_f - 1),
\end{align}
where \( \hat{\mathbf{I}}^{3D}_{n_f}=R^{-1}(\mathbf{I}_{n_f})\in\mathbb{R}^{J_{f} \times 3} \) indicates color repository re-projected from \(n_f\in N_f\) sampled images;
$\mu(\hat{\mathbf{I}}^{3D}_{j})$ indicates the mean operation across all $N_f$ views;
\(J_{f} \) denotes vertex number. Note that $J_f<J_c$, while meshes with $J_f, J_c$ vertexes are used for the fine-grained pixel space and the coarse latent space, respectively.
Variance quantifies the dispersion of data points in a dataset. 
Since the value of each vertex is accumulated from \(N_f\) pixel values, we can employ a predetermined threshold  \( \lambda \) to discern vertices that exhibit inconsistency.
We construct an indicator function as a 3D mask $\mathbf{M}\in\mathbb{R}^{J_{f} \times 1}$: 
\begin{align}
\mathbf{M}_j=\begin{cases}
1, & Var_j > \lambda\\
0, & Var_j \leq \lambda
\end{cases}.
\end{align}
After that, we perform pixel domain inpainting refinement through rendering both $\hat{\mathbf{I}}^{3D}$ and 3D mask $\mathbf{M}$ with $R(\cdot)$ into four separate views and 2D masks. Then we follow Repaint3D~\cite{repaint3D} which first
dilates inpainting mask with 8*8 kernel, and then utilize Depth-SD to inpaint them through the same prompts and depth rendered by the input mesh.
Subsequently, the inpainting procedure is carried out sequentially until all views have been inpainted and finally produces the refined texture $\tilde{\mathbf{I}}^{3D}$ as shown in the rightmost of Fig.~\ref{fig:framework}(a).

\section{Experiments}

\subsection{Experimental Settings}
\noindent\textbf{Implementation Details.}
We use the PyTorch3D library\footnote{https://github.com/facebookresearch/pytorch3d} to render 3D objects and employ Depth-SD to conduct 3D-2D collaborative denoising and inpainting refinement. 
In the collaborative denoising stage, we engage $N=9$ equidistant viewpoints with each view separated by the angular of 40$^{\circ}$.
The grid size $G_t$ is altered within two values: $0.34$ and $0.25$, while the whole volume space is normalized to $[-1, 1]$. The distance upper bound \(Z_{far}\) is set to $5$. 
Additionally, exponents \( \tau_b \),  \( \tau_w \) and \( \tau_f \) are set to $2.0$, $3.0$ and $6.0$ respectively, where \( \tau_f \) is used for pixel aggregation.
In the inpainting refinement stage, we commence by selecting $N_f=4$ view images at angular positions of $0^{\circ}, 80^{\circ}, 160^{\circ}$, and $280^{\circ}$ for the initial texture aggregation. Subsequently, a variance threshold \( \lambda \) of $0.005$ is employed to identify inconsistent areas.
For parameters related to Depth-SD, we set the denoising step as $50$ in the denoising and inpainting stages, while the 3D-2D collaborative denoising is applied to the first 45 steps.

\noindent\textbf{Datasets.}
Despite the extensive study on texture, there remains a lack of standardized datasets and evaluation metrics as a solid benchmark. 
To comprehensively evaluate the performance of the comparison methods, we construct the largest 3D shape dataset, which consists of three subsets (named \textit{SubObj}, \textit{SubShape}, and \textit{SubTex}) from three open-source 3D datasets (Objaverse~\cite{deitke2023objaverse}, ShapeNetSem~\cite{savva2015semgeo}, and ShapeNet~\cite{chang2015shapenet}). 
\textit{SubObj} is inherited from Text2Tex, containing 410 objects from Objaverse, we exclude some too-simple shapes, which remain 401 objects.
Inspired by the sampling methodology of Text2Tex, we extract 445 3D shapes across categories from the ShapeNetSem dataset to form the \textit{SubShape}, with three meshes sampled per category based on vertex count distribution.
In addition, we sample publicly authorized meshes with prompts from the Texfusion dataset to assemble 43 mesh-prompt pairs, primarily sourced from ShapeNet, into the subset \textit{SubTex}.

\noindent\textbf{Metrics.}
After reviewing relevant metrics, we employ the Fréchet Inception Distance (FID)~\cite{heusel2017gans} and the CLIP-based extension of FID, denoted as CLIP-FID~\cite{kynkaanniemi2022role}, to measure texture fidelity. Moreover, we utilize the Clip-Score~\cite{clip} metric to quantify the correspondence score between the rendered image and the textual prompt. We also utilize CLIP-Var~\cite{anonymous2023learning} to ascertain the consistency across multiple rendering views following~\cite{anonymous2023learning}. 
In detail, following the evaluation setting of Repaint3D~\cite{repaint3D}, 
we render the textured mesh from eight distinct camera perspectives, evenly spaced at 45$^{\circ}$ intervals, and overwrite the background regions with pure white color to emphasize the foreground object for more focused comparisons.
To compute the FID metric, following~\cite{repaint3D, texfusion}, ground-truth images are synthesized through Depth-SD, with each image being conditioned on the same prompt and synthesized from the designated evaluation viewpoints.
The computation of Clip-Score is achieved upon the minimum value derived from eight cosine similarity scores, each one representing the similarity between the normalized features of the CLIP image and the corresponding text features. Similarly, CLIP-Var is determined by the minimal mutual cosine similarity scores among the eight normalized CLIP image features.

\subsection{Quantitative Evaluation}
In this section, we quantitatively compare our VCD-Texture method against state-of-the-art texturing methods, including TEXTure~\cite{Texture}, Tex2Tex~\cite{chen2023text2tex}, Reapaint3D~\cite{repaint3D}, and SyncMVD~\cite{syncTexture}, which have released their source codes.
As for the comparison methods, we attain the results by running their official code with their default settings. As reported in Tab.~\ref{tab:sota_compare_table}, compared with the inpainting-based methods (Tex2Tex, TEXTure, and Reapaint3D) and the optimization-based methods (SyncMVD), our method achieves lower FID and ClipFID and higher ClipScore and ClipVar results.
The results demonstrate that our VCD-Texture method possesses superior performance in terms of both texture fidelity and multi-view consistency.
\begin{table}[!t]
\small
\caption{Quantitative comparison of different texture methods on three datasets.}
\centering
\begin{tabular}{llcccc}
\toprule
{Datasets}  & {Method} & {FID $\downarrow$} & {ClipFID $\downarrow$} & {ClipScore $\uparrow$} & {ClipVar $\uparrow$} \\ 
\midrule
\multirow{5}{*}{SubTex}  
& Texture~\cite{Texture}  & 150.21 & 26.92 & 26.90 & 82.37  \\
& Text2Tex~\cite{chen2023text2tex} &112.41 & 16.26 & 30.08 & 81.45  \\
& SyncMVD~\cite{syncTexture}  & 65.30 & 16.76 & 28.78 & 81.93  \\
& Repaint3D~\cite{repaint3D}  & 78.65  & 10.65  & 30.88   & 78.96  \\
& \textbf{VCD-Texture} & \textbf{56.29} & \textbf{6.84} & \textbf{31.65} & \textbf{83.97} \\
\midrule
\multirow{5}{*}{SubShape}  
& Texture~\cite{Texture}  & 64.78 & 18.08 & 27.20 & 82.32  \\
& Text2Tex~\cite{chen2023text2tex} & 40.46 & 7.96 & 27.76 & 82.18 \\
& SyncMVD~\cite{syncTexture}  &32.44 & 6.18 & 28.76 & 82.76  \\
& Repaint3D~\cite{repaint3D}  & 29.21 & 5.25 & 28.39 & 80.18  \\
& \textbf{VCD-Texture} & \textbf{19.46} & \textbf{2.37} & \textbf{28.98} & \textbf{82.93} \\
\midrule
\multirow{5}{*}{SubObj}  
& Texture~\cite{Texture}  & 65.30 & 16.76 & 28.78 & 81.93  \\
& Text2Tex~\cite{chen2023text2tex} & 43.71 & 7.46 & 29.27 & 82.08 \\
& SyncMVD~\cite{syncTexture}  & 34.00 & 5.60 & 30.08 & \textbf{84.52}  \\
& Repaint3D~\cite{repaint3D}  & 29.77 & 4.44 &30.30 &81.45  \\
& \textbf{VCD-Texture} & \textbf{21.19} & \textbf{2.33} & \textbf{30.42} & 83.64 \\
\bottomrule
\end{tabular}
\label{tab:sota_compare_table}
\end{table}

\subsection{Qualitative Evaluation}

\begin{figure}[tb]
  \centering
  \includegraphics[width=0.98\textwidth]{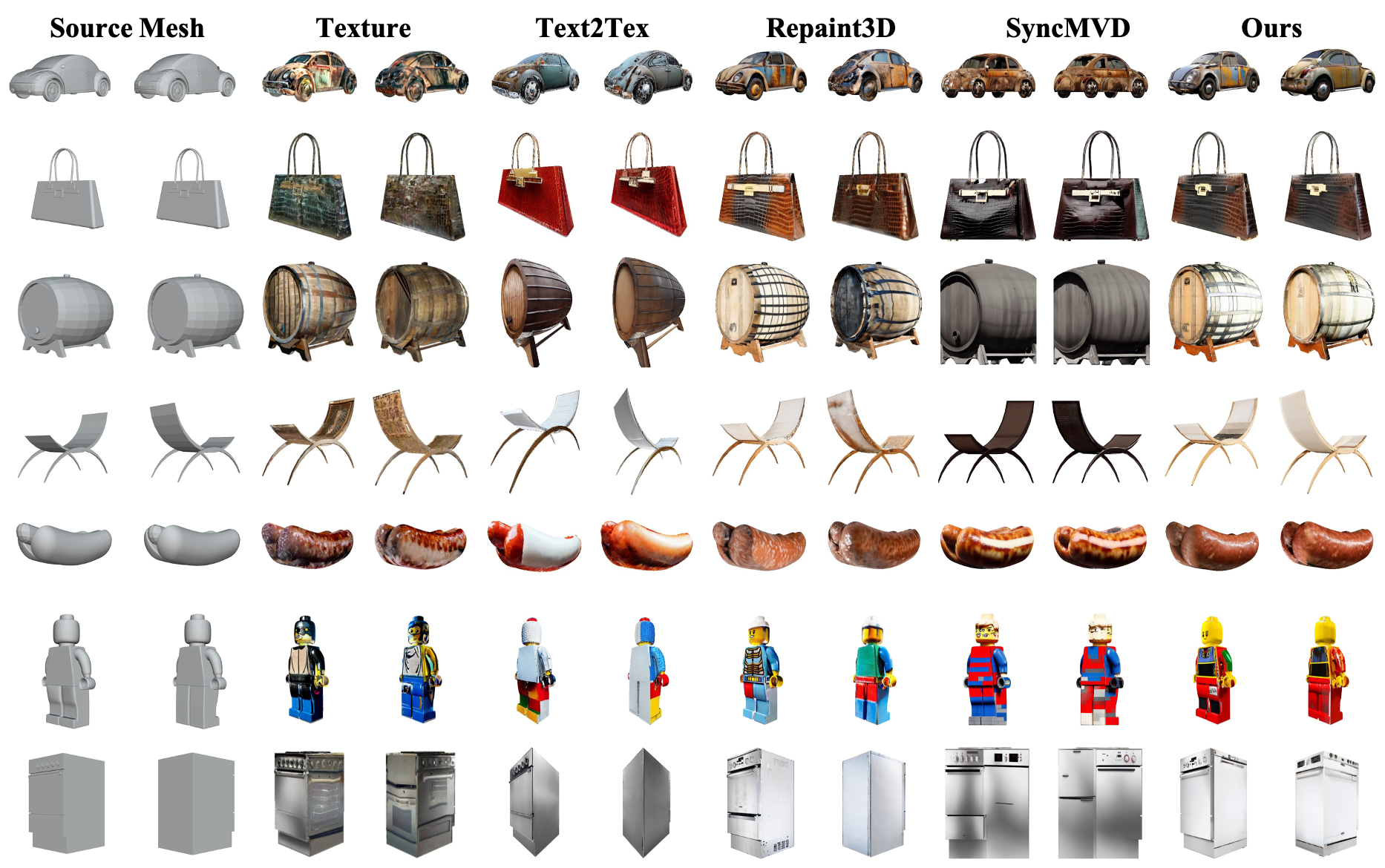}
  \caption{Qualitative comparisons of text-guided texture synthesis. Prompts from top to down are: ``old and rusty volkswagon beetle'', ``crocodile skin handbag'', ``barrel'', ``half moon chaise'', ``sausage'', ``lego'', ``electric oven''.}
  \label{fig:visual_compare}
\end{figure}
Fig.~\ref{fig:visual_compare} depicts textures generated by the same text prompt and 3D mesh. 
We can conclude that: 
1) Inpainting-based methods, Texture~\cite{Texture}, Text2Tex~\cite{chen2023text2tex}, and Repint3D~\cite{repaint3D}, tend to generate inconsistent textures on the two opposing views.
2) The optimization-based method, SyncMVD~\cite{syncTexture}, can generate more consistent textures compared to the inpainting-based ones. However, SyncMVD tends to produce textures that are excessively smooth, resulting in a loss of intricate details and fine texture quality.
2) The optimization-based method, SyncMVD~\cite{syncTexture}, can generate more consistent textures compared to the inpainting-based ones. However, SyncMVD tends to produce textures that are excessively smooth, resulting in a loss of intricate details and fine texture quality.
This deficiency could be attributed to the frequent application of synchronized multi-view diffusion coupled with the absence of a VA. Such a process generates latent features with low variance, which in turn, leads to the creation of textures that are notably smooth yet exhibit high consistency.
3) Our VCD-Texture method is capable of generating more multi-view consistent and fidelity textures compared to both the inpainting-based and optimization-based methods.

\subsection{Ablation Study}
\label{sec:abl}
\noindent\textbf{Factor-by-factor Analyzation.}
We conduct a comprehensive ablation study on the SubTex datasets to validate the effect of different components.
\begin{table}[!t]
\small
\centering
\caption{
  The ablation study for MV-AR, JNP, VA, the incorporating of distance score (DS), and inpainting refinement (IR).
}
\label{tab:ablition_study}
\vspace{-.1in}
\begin{tabular}{cccccccccc}
\toprule[1pt]
No. & MV-AR &  DS & JNP & VA & IR & {Fid $\downarrow$} & {ClipFid $\downarrow$} & {ClipScore $\uparrow$} &{ClipVar $\uparrow$}   \\ \midrule
(1) & $\checkmark$ & & & & & 58.87 & 7.39 & 31.32 & 82.87 \\
(2) & $\checkmark$ & $\checkmark$ & & & & 58.73 & 7.25 & 31.47 & 83.12 \\    
(3) & $\checkmark$ & $\checkmark$ & $\checkmark$ & & & 58.02 & 7.16 & 31.57 & 83.62\\
(4) & $\checkmark$ & $\checkmark$ & $\checkmark$ & $\checkmark$ & & \textbf{56.05} & 6.88 & 31.62 & 83.94 \\
(5) & $\checkmark$ & $\checkmark$ & $\checkmark$ & $\checkmark$ &$\checkmark$  & 56.29 & \textbf{6.84} & \textbf{31.65} & \textbf{83.97}  \\
\bottomrule[1pt]
\end{tabular}
\end{table}
The experimental results are summarized in Tab.~\ref{tab:ablition_study}, from which we can draw the following conclusions:
\begin{itemize}
\item Comparing the models (1) and (2), the distance score (DS) improves Clip-FID and multi-view consistency. This validates pixel distance is a valid metric that can aid the view score in selecting more consistent pixels in the aggregating process.
\item Comparing the models (2) and (3), 
the integration of JNP facilitates the exchange of information within the 3D domain, which improves the semantic metric ClipScore and also enhances the consistency across views.
\item Comparing models (3) and (4), incorporation of VA leads to a significant improvement in the FID metric, this proves that rectifying feature variance to align original diffusion distribution after rasterization helps to generate high-fidelity images. 
\item Comparing models (4) and (5), since the inconsistent regions are small, thus inpainting refinement doesn't contribute much more improvements in evaluation metrics. However, the subsequent visual comparative analysis clearly reveals that inpainting refinement is instrumental in mitigating blurred regions and rectifying visual artifacts.
\end{itemize}

\begin{figure}[!t]
  \centering
  \includegraphics[width=1.0\textwidth]{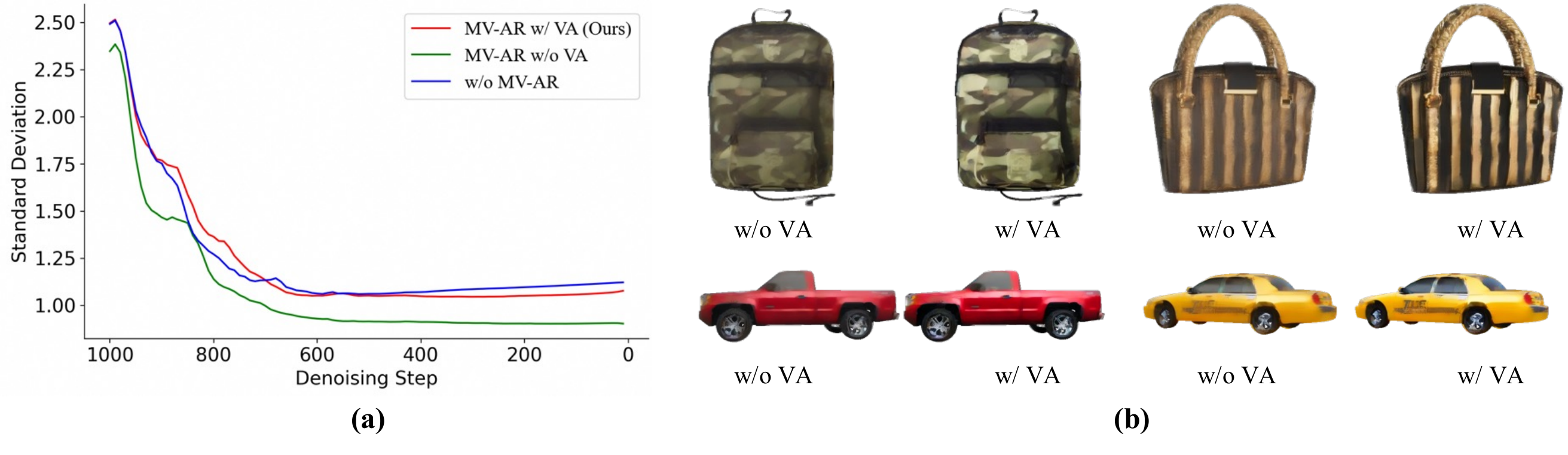}
  \vspace{-.15in}
  \caption{The effectiveness of VA. (a) shows the standard deviation curve of three denoising policies; (b) showcases the qualitative comparison with and without (w.o) VA.}
  \label{fig:variance}
  \vspace{-0.15in}
\end{figure}

\noindent\textbf{Variance Alignment Analysis.}
To elucidate the process of variance reduction, 
we present the standard deviation trajectories of latent features during the denoising phase in Fig.~\ref{fig:variance}.
The blue curve (without MV-AR) depicts the denoising process without consistent multi-view projection.
The green curve (MV-AR without VA) illustrates the feature variance resulting from a consistent denoising projection, \emph{i.e.}, lifting the feature to 3D space and rasterizing it back to 2D without VA.
The red curve (MV-AR with VA) represents the feature variance when VA is incorporated into the MV-AR.

\begin{figure}[!t]
  \centering
  \includegraphics[width=1.0\textwidth]{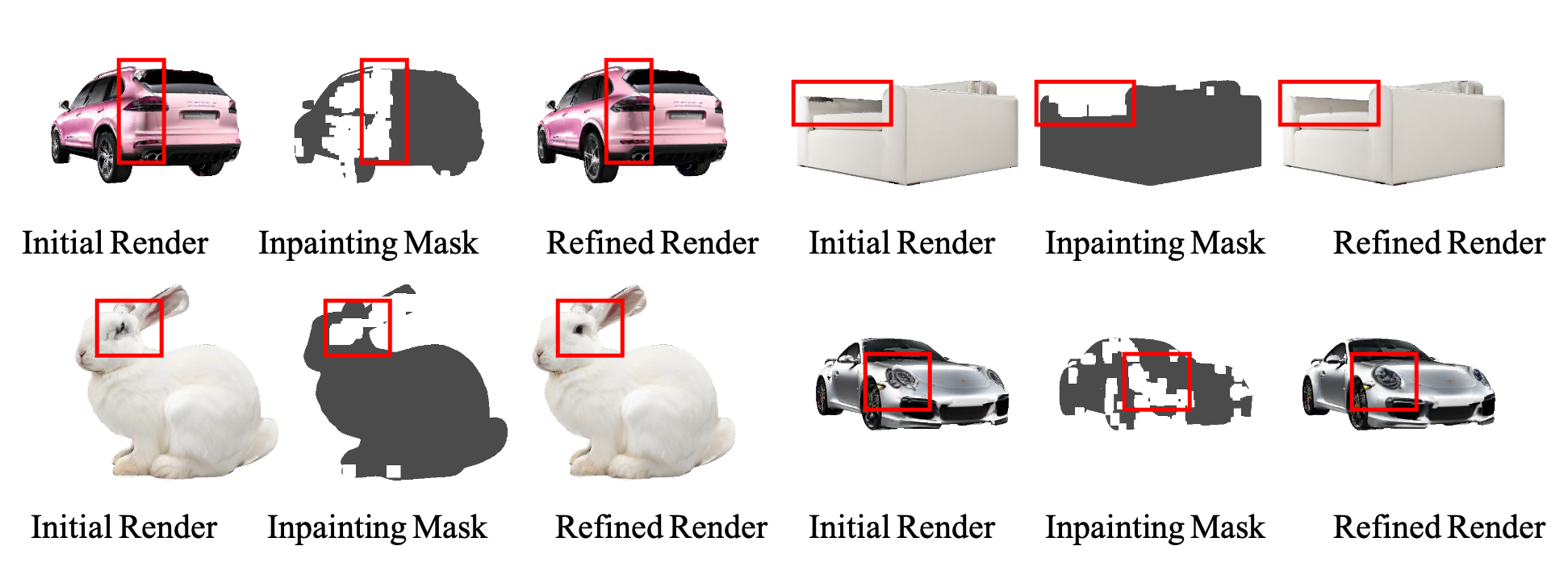}
  \vspace{-.15in}
  \caption{
  Qualitative results of the inpainting refinement.}
  \label{fig:inpchange}
  \vspace{-.15in}
\end{figure}

A comparative analysis of these curves reveals that the green curve consistently maintains a lower trajectory than the standard blue line, indicative of the variance diminishing process and congruent with the theoretical proof's conclusions. The trajectory of the red curve (MV-AR with VA), which adopts VA, displays a similar pattern to the blue curve and ultimately converges just below the blue line, which validates the efficacy of the proposed VA method. The lower endpoint of the red line can be attributed to the features generated by consistent projection being more consistent than those in the general denoising process.

\noindent\textbf{Effectiveness of Inpainting Refinement.}
As illustrated in Fig.~\ref{fig:inpchange}, we visualize the texture improvements w.r.t. the inpainting refinement process on four examples of \textit{SubTex} dataset. We can observe that the inpainting mask can accurately identify those inconsistent pixels, and the subsequent pixel domain inpainting refinement is capable of refining the inconsistent regions and achieving more fidelity and high-quality texturing results.


\section{Conclusion}
We propose a novel collaborative denoising 3D texture synthesis approach, VCD-Texture, to mitigate the gap between the 2D diffusion generation and 3D objects. 
VCD-Texture injects 3D geometries into the 2D diffusion denoising process, designing Joint Noise Prediction (JNP) and Multi-View Aggregation-and-Rasterization (MV-AR) modules to incorporate features in 2D and 3D space. 
Moreover, we theoretically analyze the variance bias issues caused by the rasterization in MV-AR, which is eliminated by the proposed Variance Alignment (VA) technique.  
To further reduce the intrinsic discrepancy in latent feature and image pixel domains, we design an inpainting refinement to rectify identified inconsistent regions.
Through a collaborative denoising process and inpainting refinement process, VCD-Texture enables the generation of consistent textures with diverse and high-quality details.

%
%
\section{Supplementary Materials}
This supplementary material first presents \textit{Additional Results}, which are organized into five sections: 1) trade-off between consistency and time, 2) time cost comparisons, 3) visual ablation studies of variance alignment, 4) visual comparisons with TexFusion, and 5) more visual comparison results. 
Next, the \textit{limitations} of the proposed method are discussed. 
Thereafter, we offer \textit{Algorithm Details} about the core JNP and MV-AR modules.
Fourthly, we deduce the inequality of \textit{Rasterization Variance Reduction}.
Finally, further details about the \textit{evaluation dataset} are provided.

\subsection{Additional Results}
\subsubsection{Trade-off between Consistency, Fidelity and Time.}
Our experiments reveal that the number of views significantly impacts efficiency and performance. We conduct ablation studies to analyze the influence of view numbers. Tab.~\ref{tab:ablation_view} summarizes the results, from which three key conclusions can be drawn:
\begin{itemize}
\item When comparing models (1) to (4), an increasing number of views results in an opposite trend between fidelity metrics (FID, ClipFID) and the consistency metric (ClipVar). More views cause more overlaps, leading to higher consistency; however, more overlap areas cause more blurriness, resulting in lower fidelity.
\item Comparing models (1) to (4), the number of views and the cost time are positively correlated, more views require more generation time. Additionally, by comparing models (5) and (6), the inpainting refinement stage takes approximately 11 seconds.

\item Comparing models (4) and (5), we propose a sampling view policy that utilizes 9 views during the denoising process and samples a subset of 4 views during the pixel aggregation process. This trade-off between consistency, fidelity, and time enables achieving high consistency while maintaining relatively higher fidelity.
\end{itemize}

\begin{table}[]
\caption{Ablations for view number on \textit{SuxTex} dataset. LView represents latent views used in the denoising process, $4 \times 90$ means $4$ views with a $90$-degree interval each, PVnum denotes the view number used for pixel aggregation.}
\centering
\begin{tabular}{c|c|c|c|cccc|c}
\toprule
No. &{LView}  & {PVnum} & Inpaint & {FID $\downarrow$} & {ClipFID $\downarrow$} & {ClipScore $\uparrow$} & {ClipVar $\uparrow$} & {Runtime (s) $\downarrow$} \\ 
\midrule  
(1) &$4 \times 90$& $4$& & \textbf{51.47} & \textbf{6.00} & 31.54 & 82.13 & \textbf{70.6} \\
(2) &$6 \times 60$& $6$& &55.32 & 6.77 & \textbf{31.81} & 83.11 & 93.4  \\
(3) &$8 \times 45$& $8$& &57.17 & 7.01 & 31.57 & 83.36 & 98.5  \\
(4) &$9 \times 40$& $9$& &59.71 & 7.56 & 31.69 & \textbf{84.03} & 103.4  \\
(5) &$9 \times 40$& $4$& &56.05 & 6.88 & 31.62 & 83.94 & 93.1 \\
(6) &$9 \times 40$& $4$&$\checkmark$&56.29 & 6.84 & 31.65 & 83.97 & 104.0 \\
\bottomrule
\end{tabular}
\label{tab:ablation_view}
\end{table}

\subsubsection{Time Cost Comparison.}
We compared the runtime efficiency of VCD-Texture against prior approaches on a GPU server with eight NVIDIA RTX A800 GPUs.
The results are presented in Tab.~\ref{tab:time_compare}.
For clarity, we organized the evaluated methods into two categories: Fitting (training-based neural networks for multi-view texture assimilation) and Re-Projection (rasterization-based re-projection without training). Re-Projection methods significantly outperformed Fitting methods, being an order of magnitude faster. Within Re-Projection, Texture~\cite{Texture} demonstrates the quickest performance times under default configuration, which were on par with SyncMVD~\cite{syncTexture}.

Since we apply a fine mesh $M_f$ in pixel aggregation, which takes more time in reprojecting colors to the mesh, and also introduce an additional inpainting refinement stage, this results in a marginal increase in runtime compared to Texture and SyncMVD~\cite{syncTexture}. 
To improve efficiency, we design a fast version of our algorithm (Ours-Fast), which incorporates three optimizations: 1) reducing denoising steps from 50 to 30; 2) remeshing the source mesh at a coarse level (from 256 to 128 resolution) to speed up color reprojection; and 3) removing the inpainting stage. This fast version achieves the fastest speed while maintaining high performance.

\begin{table}[]
\centering
\caption{Runtime  Comparisons. VNum denotes the number of views, TType represents texture drawing type, and Re-Proj means texture drawing by color re-projection. `Ours-Fast' is implemented with fewer denoising steps from 50 to 30, lower mesh resolution from 256 to 128, and without the inpainting stage, which still outperforms other competitors.}
\label{tab:time_compare}
\begin{tabular}{l|c|c|cccc|c}
\toprule[1pt]
Method & TType & VNum & {FID $\downarrow$} & {ClipFID $\downarrow$} & {ClipScore $\uparrow$} & {ClipVar $\uparrow$}& Runtime (s) $\downarrow$ \\
\midrule
Text2Tex~\cite{chen2023text2tex} & Fitting & 36 &112.41 & 16.26 & 30.08 & 81.45 & 842.20 \\
Repaint3D~\cite{repaint3D} & Fitting & 9& 78.65  & 10.65  & 30.88   & 78.96 & 611.60  \\ 
Texture~\cite{Texture} & Re-Proj & 8& 150.21 & 26.92 & 26.90 & 82.37 & 79.50  \\ 
SyncMVD~\cite{syncTexture} & Re-Proj & 10 & 65.30 & 16.76 & 28.78 & 81.93 &  83.30 \\ 
Ours & Re-Proj & 9 &\textbf{56.29} & \textbf{6.84} & \textbf{31.65} & \textbf{83.97} & 104.00 \\ 
Ours-Fast & Re-Proj & 9 & 62.57 & 9.73 & 31.42 & 83.10 & \textbf{74.40} \\
\bottomrule[1pt]
\end{tabular}
\end{table}

\subsubsection{Visual Ablation of Variance Alignment.}
The analysis of the standard deviation curve, presented in Fig.~\ref{fig:variance}(a), indicates that iterative rasterization within the denoising phase results in a reduction in the magnitude of feature variance, which is likely to cause Depth-SD to produce images that are blurred or excessively smooth. To validate this conclusion, we ran experiments with and without applying variance alignment under 100 denoising steps, respectively. Fig.~\ref{fig:variance}(b) shows the generated images. Comparing images without VA (first and third columns) to images with VA (second and fourth columns), we can observe that images without VA have lower contrast and much more blurriness, in contrast, images with VA exhibit higher fidelity and more distinct textural details. This proves that our proposed variance alignment can improve the image quality produced by Depth-SD when frequently conducting rasterization.

\subsubsection{Visual Comparison with TexFusion.}
\begin{figure}[!t]
  \centering
  \includegraphics[width=0.98\textwidth]{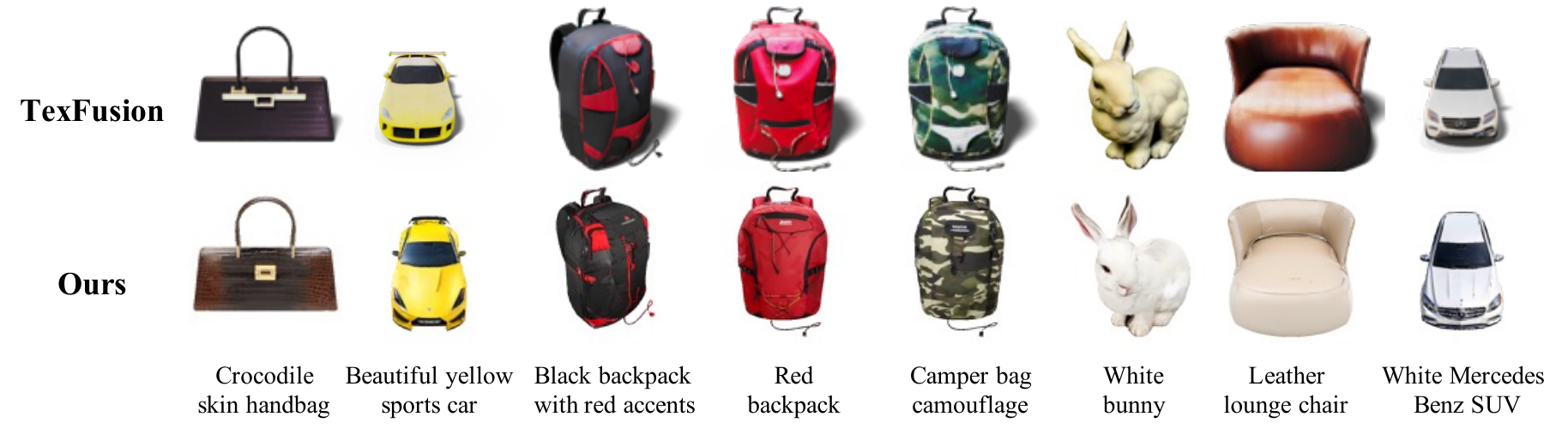}
  \caption{Qualitative comparisons with TexFusion.}
  \label{fig:sup_texfusion_compare}
\end{figure}

TexFusion~\cite{texfusion} employs a similar latent texture methodology. However, at the time of writing this manuscript, the authors had not made their code or textured mesh data publicly available. Consequently, we have opted to utilize the images presented in the original publication for visual comparison. To visualize the top view, we have incorporated two additional views (46$^{\circ}$, 0$^{\circ}$) and (46$^{\circ}$, 180$^{\circ}$) into our default horizontal viewpoints.
Fig.~\ref{fig:sup_texfusion_compare} presents a visual comparison, from which we observed higher fidelity and finer details in our results.

While TexFusion~\cite{texfusion} integrates an ancillary VGG-based loss to diminish the discrepancy between the latent and pixel domains. Nevertheless, it fails to effectively address the issue of pixel inconsistencies across views, resulting in textures that are perceptibly blurred.
In contrast, our approach employs a two-stage pipeline that initially enforces consistent feature generation within the latent domain, followed by a refinement process through pixel domain inpainting. This two-stage strategy synergistically enhances the consistency and fidelity of the synthesized textures.

\subsubsection{More Visual Comparison.}
\begin{figure}[!t]
  \centering
  \includegraphics[width=0.98\textwidth]{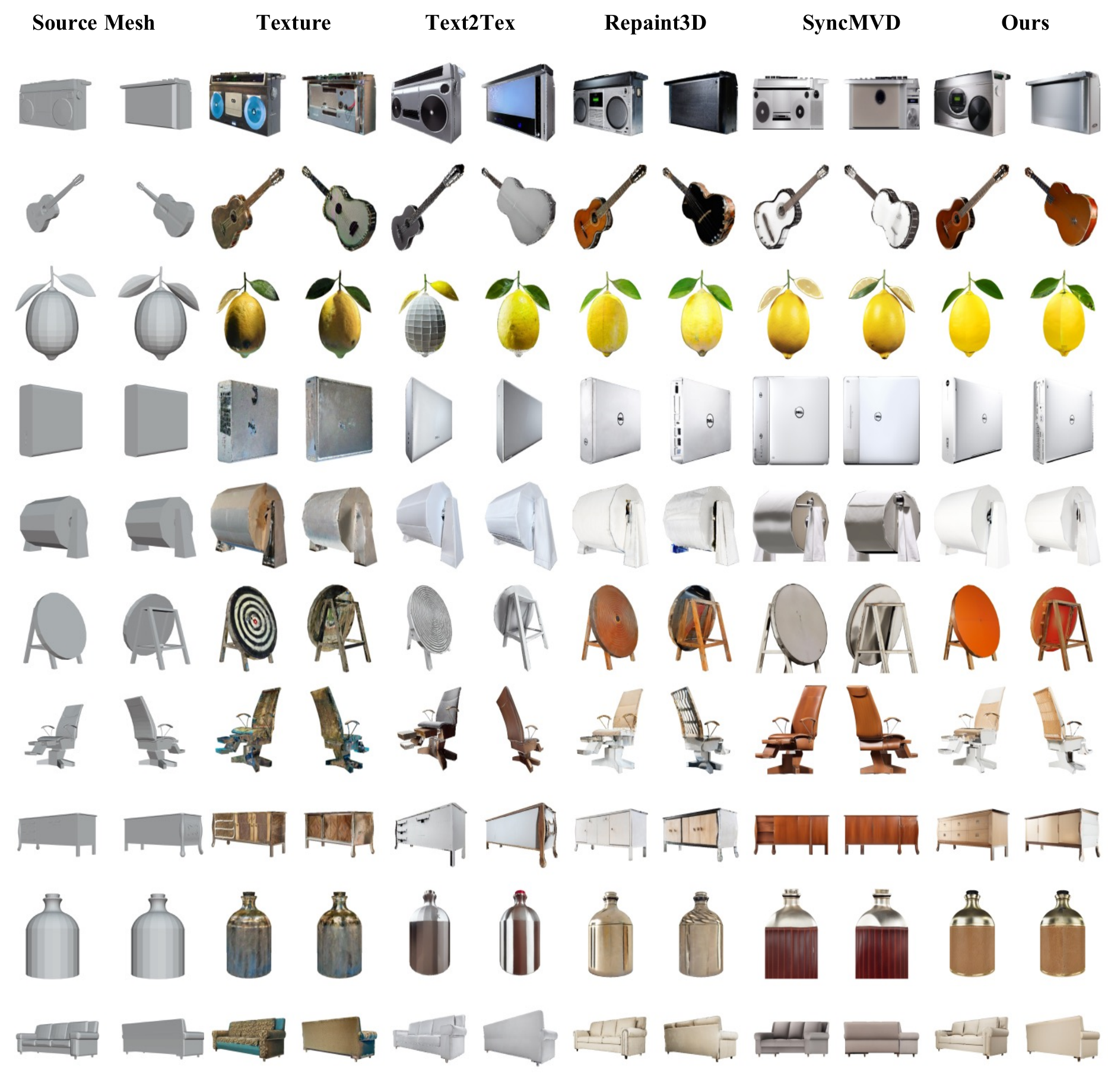}
  \caption{Qualitative comparisons of text-guided texture synthesis. Prompts from top to down are: ``CD player'', ``banjo'', ``lemon'', ``dell inspiron white'', ``Toilet paper holder with tp'', ``target'', ``Chair stool armchair stuhl'', ``kare sideboard janus'', ``flask'' , ``sofa''.}
  \label{fig:sup_sota_compare}
\end{figure}

As depicted in Fig.~\ref{fig:sup_sota_compare}, we provide more qualitative comparisons against state-of-the-art counterparts, further validating the effectiveness and superiority of our proposed approach.

\subsection{Limitations}
Our research is subject to two principal limitations, primarily attributable to the constraints inherent in the pre-trained diffusion model. Firstly, the issue of pre-illumination: the images synthesized by the SD model display variations in luminance, leading to instances of local overexposure in textures. This challenge has been addressed by the development of a diffusion model~\cite{paint3d} devoid of lighting effects. Secondly, we identify the presence of local artifacts: the Depth-SD technique synthesizes images based on a combination of depth maps and textual prompts. Due to the discrepancy in complexity between the rudimentary 3D mesh geometry and the intricate training images, the resulting depth map of the synthesized image fails to align precisely with the conditional depth map. This misalignment results in the projection of unmatched pixel colors onto the corresponding 3D vertices, thereby generating local artifacts. A potential solution to this issue lies in fine-tuning texture using methods such as PatchGAN~\cite{patchgan}, which employs a contrastive approach to learning the distribution of image patches.

\subsection{Dataset and Evaluation details}
We evaluated our method on three datasets, the statistical details of which are presented in Table~\ref{tab:datasets}. Notably, there are some meshes listed in TexFusion~\cite{texfusion} were not found in the ShapeNet~\cite{chang2015shapenet} dataset, and we utilize meshes from the same category as replacements for those invalid meshes. Since the Fréchet Inception Distance (FID) metric can be influenced by the number of evaluation images, we report the ground-truth numbers (GT-Num) in the third column of Table~\ref{tab:datasets}. 

The comprehensive list of mesh names and prompts employed for each dataset is provided in the supplementary material, whose file names are: \textit{subtex.txt}, \textit{subobj.txt} and \textit{subshape.txt}. 
We use the same data loader as Repaint3D~\cite{repaint3D}, which can be found at: \href{https://github.com/kongdai123/repainting_3d_assets/blob/256af2cbeceaa009cf67fa424de85a161355c4b7/repainting_3d_assets/main_shapenet.py#L11}{Data Loader of Repaint3D}

\begin{table}[]
\centering
\caption{Evaluation Datsets.}
\label{tab:datasets}
\begin{tabular}{c|c|c}
\toprule[1pt]
Name & Num & GT-Num  \\
\midrule
\textit{SubTex} & 43 & 344 \\
\textit{SubShape} & 445 & 3560  \\ 
\textit{SubObj} & 401 & 3208  \\ 
\bottomrule[1pt]
\end{tabular}
\end{table}

\vspace{-.20in}
\subsection{Proof of Rasterization Variance Reduction}
Formally, Jensen's inequality states that if $\varphi(\cdot)$ is a convex function, and \( z_{i\in N_z} \) are points in interval \( Z \), where \(N_z\) is the number of sampling points, then for any non-negative weights \( \lambda_i\) that satisfy the condition: \( \sum^{N_z}_{i=1} \lambda_i = 1 \), the following inequality holds:
\begin{align}
    \varphi\left(\sum_{i=1}^{N_z} \lambda_i z_i\right) \leq \sum_{i=1}^{N_z} \lambda_i \varphi(z_i).
    \label{eq:jensen_inequality}
\end{align}

For random variable set $\mathbf{X_i}$, the variable \(\mathbf{Y_j}\) are combined by variables sampled from $\mathbf{X_i}$, which is computed by:
\begin{align}
    \mathbf{Y_j} = \sum^{N}_{i=1} \lambda_i \cdot \mathbf{X_i}_j
    \label{eq:convex_combination}
\end{align}
where \( j\) denotes index in \(\mathbf{Y}\), \(i\) is the variable set index.  \(\lambda_i\) represents non-negative weights, which satisfy the condition \( \sum^{N}_{i=1} \lambda_i = 1, \lambda_i \geq 0 \). 
Let $M$ denotes the element number of \(\mathbf{Y}\), The variance of \(\mathbf{Y}\) is defined by:
\begin{align}
    Var(\mathbf{Y}) &= \sum^{M}_{j=1} [\mathbf{Y}_j - \mu(\mathbf{Y})]^2 / (M - 1) \\
    &= \sum^{M}_{j=1} [\sum^{N}_{i=1} \lambda_i \cdot \mathbf{X_i}_j - \sum^{N}_{j=1} \lambda_i \cdot \mu (\mathbf{X_i})]^2 / (M - 1) \\
    &= \sum^{M}_{j=1}  [\sum^{N}_{i=1} \lambda_i \cdot (\mathbf{X_i}_j - \mu(\mathbf{X_i})]^2 / (M - 1) \\
    \label{eq:convex_combination}
\end{align}
As square is a convex function, based on the Jensen's inequality \ref{eq:jensen_inequality}, For each variable \(\mathbf{Y}_j\),  we have:

\begin{align}
  [\sum^{N}_{i=1} \lambda_i \cdot (\mathbf{X_i}_j - \mu(\mathbf{X_i})]^2 
 \leq
 \lambda_i \cdot \sum^{N}_{i=1} [\mathbf{X_i}_j - \mu(\mathbf{X_i})]^2
    \label{eq:convex_combination}
\end{align}

 Let \(E_{Y|j}\) and \( E_{X|ij} \) denote the squared deviation term in \(\mathbf{Y}\) and \(\mathbf{X_i}\) separately, which are defined as  \(E_{Y|j} = [\mathbf{Y}_j - \mu(\mathbf{Y})]^2\), \( E_{\mathbf{X_i}|j} =  [\mathbf{X_i}_j - \mu(\mathbf{X_i})]^2 \). Referring previous inequality, we have:

\begin{align}
E_{Y|j} \leq \lambda_i \cdot \sum^{N}_{i=1} E_{\mathbf{X_i}|j}
\label{eq:convex_combination}
\end{align}
This means each squared deviation term \(E_{Y|j}\) in \(\mathbf{Y}\) is no large than the linear combined squared deviation term \( E_{\mathbf{X_i}|j} \) in \(\mathbf{X_i}\) . And then apply the expectation with total number \(M\), we have:

\begin{align}
Var(\mathbf{Y}) &= \sum^{M}_{j=1} E_{Y|j} / (M - 1) \\
&\leq \sum^{M}_{j=1} \lambda_i \cdot \sum^{N}_{i=1} E_{\mathbf{X_i}|j} / (M - 1) = \sum^{N}_{i=1} \lambda_i \cdot \sum^{M}_{j=1} E_{\mathbf{X_i}|j} / (M - 1)
\end{align}
As \( Var(\mathbf{X}_i) =\sum^{M}_{j=1} E_{\mathbf{X_i}|j} / (M - 1) \), thus we have:
\begin{align}
    Var(\mathbf{Y}) \leq \sum^{N}_{i=1} \lambda_i \cdot Var(\mathbf{X}_i)
\end{align}

\subsection{Algorithm Details}

We present pseudo codes for two core modules: Joint Noise Prediction in Algorithm~\ref{alg:JNP},  Multi-View Aggregation and Rasterization (MV-AR) in Algorithm~\ref{alg:mv-ar}.
To optimize efficiency, the Joint Noise Prediction module was solely implemented at the highest resolution of the U-Net architecture. Additionally, cross-attention mechanisms operating in 3D space were also attempted but did not yield performance improvements.

\begin{algorithm}
\caption{Joint Noise Prediction Algorithm}
\label{alg:JNP}
\begin{algorithmic}[1]
\Statex \textbf{Input}: Coarse mesh $M_c$, Cameras $C_n$, denoised latent feature $\mathbf{F}_{n}^{2D}$
\Statex \textbf{Parameters}: View number $N$, Vertex face index $\{f_u\}^3_{u=1}$ in each face, Vertex Coordinate $P^v_{j}$, Feature size $(w, h)$, Rasterizaiton function $R(n)$. In Pytorch3D library, $R(n)$ contain three output tensors: Depth map tensor ${\hat{D}_n}$, barycentric coordinate tensor $B_n$ and pixel and mesh face relation tensor $R^p_n$ specifying the indices of the faces which overlap each pixel.
\Statex \textbf{Return}: $\tilde{\mathbf{F}}_{l|n}^{2D}$: updated latent features at level $l$ of U-Net
\Procedure{Joint Noise Prediction Algorithm}{}
\For{each level $l \in L$}
\Statex 
    \State \textbf{\#Step1}: Extract 3D features $\mathbf{F}_l^{3D}$
    \For{each view $n \in N$}
        \State Build rasterization relation $R(n)$ =  \(Pytorch3D.Render(M_c, C_n)\)
        \State Compute 3D coordinates $P^F_{n,i} = \sum_{u=1}^{3} ({B^{u}_{n, i}})^2 \cdot P^v_{f_u}$ 
        \State Extract $\mathbf{F}_l^{3D}$ from 2D foreground features with 3D coordinates.
    \EndFor
\Statex 
    \State \textbf{\#Step2}: Split 3D features into groups
    \State Compute bounding box $B^p$ of 3D space features $\mathbf{F}_l^{3D}$.
    \State Group $\mathbf{F}_l^{3D}$ into different groups $\mathbf{F}_{l|g}^{3D}$ by grid size $G_t$
\Statex 
    \State \textbf{\#Step3}: Compute view-aware 2D self-attention in each 2D plane
    \For{each view index $n \in N$}
        \State Compute $\tilde{\mathbf{F}}_{l|n}^{2D}$ = \(SelfAttn(\mathbf{F}_{l|n}^{2D})\)
    \EndFor
\Statex 
    \State \textbf{\#Step4}:  Compute grid-aware 3D self-attention in each 3D grid
    \For{each group index $g \in G^t$}
        \State Compute $\tilde{\mathbf{F}}_{l|g}^{3D}$ = \(SelfAttn(\mathbf{F}_{l|g}^{3D})\)
    \EndFor
    \State Obtain 3D features $\tilde{\mathbf{F}}_{l|n}^{3D}$ in 2D space by removing coordinates of $\tilde{\mathbf{F}}_l^{3D}$
\Statex 
    \State \textbf{\#Step5}: Combine features from 2D and 3D space
    \State $\tilde{\mathbf{F}}_{l|n}^{2D}$ = \(Mean(\tilde{\mathbf{F}}_{l|n}^{3D} + \tilde{\mathbf{F}}_{l|n}^{2D})\)
\EndFor
\EndProcedure
\end{algorithmic}
\end{algorithm}

\begin{algorithm}
\caption{Multi-View Aggregation and Rasterization}
\label{alg:mv-ar}
\begin{algorithmic}[1]
\Statex \textbf{Input}: Coarse mesh $M_c$, Cameras $C_n$, denoised latent feature $\mathbf{F}_{n}^{2D}$
\Statex \textbf{Parameters}: View number $N$, Vertex index $\{j\}^{J_c}_{j=1}$, Feature size $(w, h)$, Upper bound of scene distance $Z_{\text{far}}$, Exponents $\tau_b, \tau_d, \tau_w$ for the power function, Rasterizaiton function $R(n)$, In Pytorch3D library, $R(n)$ contain three output tensors: Depth map tensor ${\hat{D}_n}$, barycentric coordinate tensor $B_n$ and pixel and mesh face relation tensor $R^p_n$ specifying the indices of the faces which overlap each pixel.

\Statex \textbf{Return}: $\tilde{\mathbf{X}}_{n}^{2D}$: updated latent predictions for each view $n \in N$
\Procedure{Multi-View Aggregation and Rasterization}{}
\Statex 
\State \textbf{\#Step1}: Initialize view scores $S_n$ and distance scores $D_n$ for each view $n \in N$
\For{each view $n \in N$}
    \State Build rasterization relation $R(n)$ =  \(Pytorch3D.Render(M_c, C_n)\)
    \State Compute view score $S_n$ = \(Cosine(Normal(M_c), Direction(C_n)) \)
    \State Compute depth score \( D_n = 1 - \hat{D}_n / Z_{\text{far}} \)
\EndFor
\Statex 
\State \textbf{\#Step2}:Initialize vertex features $\hat{\mathbf{X}}_{n,j}$ and vertex weights $W_{n,j}$
\For{each view $n \in N$ and vertex index $j \in J_c$}
    \State Compute normalization factor $\eta = \sum_i \psi(B_{n,i}, \tau_b)$
    \State Re-project 2D to 3D $\hat{\mathbf{X}}_{n,j}^{3D} = \sum_i X_{n,i}^{2D} \cdot \psi(B_{n,i}, \tau_b) / \eta$, the relation of each vertex $\hat{\mathbf{X}}_{n,j}^{3D}$ and 2D pixel values $\mathbf{X}_{n,i}^{2D}$ are derived from $R^p_n$.
    \State Compute view weight $W_{n,j} = \sum_i S_{n,i} \cdot \psi(D_{n,i}, \tau_d) \cdot \psi(B_{n,i}, \tau_b) / \eta$
\EndFor

\Statex 
\State \textbf{\#Step3}:Aggregate each view features to final texture feature
\For{each vertex index $j \in J_c$}
    \State Compute normalization factor $\omega = \sum_n \psi(W_{n,j}, \tau_w)$
    \State View Aggregation $\hat{\mathbf{X}}^{3D}_{j} = \sum_n \hat{\mathbf{X}}_{n,j}^{3D} \cdot \psi(W_{n,j}, \tau_w) / \omega$
\EndFor
\Statex
\State \textbf{\#Step4}:Rasterize final texture feature to 2D plane feature 
\For{each view $n \in N$}
    \State Compute $\tilde{\mathbf{X}}_{n}^{2D}$ =  \(Pytorch3D.Render(M_c, C_n, \hat{\mathbf{X}}^{3D})\)
    \State Replace background features in $\tilde{\mathbf{X}}_{n}^{2D}$ with $\mathbf{X}_{n}^{2D}$ 
\EndFor
\EndProcedure
\end{algorithmic}
\end{algorithm}

\clearpage
\bibliographystyle{splncs04}
\bibliography{main}

\begin{thebibliography}{10}
\providecommand{\url}[1]{\texttt{#1}}
\providecommand{\urlprefix}{URL }
\providecommand{\doi}[1]{https://doi.org/#1}

\bibitem{anonymous2023learning}
Anonymous: Learning pseudo 3d guidance for view-consistent 3d texturing with 2d diffusion (2023), \url{https://openreview.net/forum?id=2A199SAhW3}

\bibitem{balas2006texture}
Balas, B.J.: Texture synthesis and perception: Using computational models to study texture representations in the human visual system. Vision research  \textbf{46}(3),  299--309 (2006)

\bibitem{texfusion}
Cao, T., Kreis, K., Fidler, S., Sharp, N., Yin, K.: Texfusion: Synthesizing 3d textures with text-guided image diffusion models. In: Proceedings of the IEEE/CVF International Conference on Computer Vision. pp. 4169--4181 (2023)

\bibitem{chang2015shapenet}
Chang, A.X., Funkhouser, T., Guibas, L., Hanrahan, P., Huang, Q., Li, Z., Savarese, S., Savva, M., Song, S., Su, H., et~al.: Shapenet: An information-rich 3d model repository. arXiv preprint arXiv:1512.03012  (2015)

\bibitem{chen2023text2tex}
Chen, D.Z., Siddiqui, Y., Lee, H.Y., Tulyakov, S., Nie{\ss}ner, M.: Text2tex: Text-driven texture synthesis via diffusion models. arXiv preprint arXiv:2303.11396  (2023)

\bibitem{chen2004texture}
Chen, Y., Ip, H.H.: Texture evolution: 3d texture synthesis from single 2d growable texture pattern. The Visual Computer  \textbf{20},  650--664 (2004)

\bibitem{chen20073d}
Chen, Y.: 3d texture mapping for rapid manufacturing. Computer-Aided Design and Applications  \textbf{4}(6),  761--771 (2007)

\bibitem{tango}
Chen, Y., Chen, R., Lei, J., Zhang, Y., Jia, K.: Tango: Text-driven photorealistic and robust 3d stylization via lighting decomposition. Advances in Neural Information Processing Systems  \textbf{35},  30923--30936 (2022)

\bibitem{chen2022auv}
Chen, Z., Yin, K., Fidler, S.: Auv-net: Learning aligned uv maps for texture transfer and synthesis. In: Proceedings of the IEEE/CVF Conference on Computer Vision and Pattern Recognition. pp. 1465--1474 (2022)

\bibitem{cula20043d}
Cula, O.G., Dana, K.J.: 3d texture recognition using bidirectional feature histograms. International Journal of Computer Vision  \textbf{59},  33--60 (2004)

\bibitem{deitke2023objaverse}
Deitke, M., Schwenk, D., Salvador, J., Weihs, L., Michel, O., VanderBilt, E., Schmidt, L., Ehsani, K., Kembhavi, A., Farhadi, A.: Objaverse: A universe of annotated 3d objects. In: Proceedings of the IEEE/CVF Conference on Computer Vision and Pattern Recognition. pp. 13142--13153 (2023)

\bibitem{gan}
Goodfellow, I., Pouget-Abadie, J., Mirza, M., Xu, B., Warde-Farley, D., Ozair, S., Courville, A., Bengio, Y.: Generative adversarial networks. Communications of the ACM  \textbf{63}(11),  139--144 (2020)

\bibitem{heusel2017gans}
Heusel, M., Ramsauer, H., Unterthiner, T., Nessler, B., Hochreiter, S.: Gans trained by a two time-scale update rule converge to a local nash equilibrium. Advances in neural information processing systems  \textbf{30} (2017)

\bibitem{ddpm}
Ho, J., Jain, A., Abbeel, P.: Denoising diffusion probabilistic models. Advances in neural information processing systems  \textbf{33},  6840--6851 (2020)

\bibitem{patchgan}
Isola, P., Zhu, J.Y., Zhou, T., Efros, A.A.: Image-to-image translation with conditional adversarial networks. In: Proceedings of the IEEE conference on computer vision and pattern recognition. pp. 1125--1134 (2017)

\bibitem{jetchev2016texture}
Jetchev, N., Bergmann, U., Vollgraf, R.: Texture synthesis with spatial generative adversarial networks. arXiv preprint arXiv:1611.08207  (2016)

\bibitem{stylegan}
Karras, T., Laine, S., Aila, T.: A style-based generator architecture for generative adversarial networks. In: Proceedings of the IEEE/CVF conference on computer vision and pattern recognition. pp. 4401--4410 (2019)

\bibitem{klodt2018supervising}
Klodt, M., Vedaldi, A.: Supervising the new with the old: learning sfm from sfm. In: Proceedings of the European conference on computer vision (ECCV). pp. 698--713 (2018)

\bibitem{kniaz2018thermal}
Kniaz, V., Mizginov, V.: Thermal texture generation and 3d model reconstruction using sfm and gan. The International Archives of the Photogrammetry, Remote Sensing and Spatial Information Sciences  \textbf{42},  519--524 (2018)

\bibitem{kniss2005octree}
Kniss, J., Lefohn, A., Strzodka, R., Sengupta, S., Owens, J.D.: Octree textures on graphics hardware. In: ACM SIGGRAPH 2005 Sketches, pp. 16--es (2005)

\bibitem{kry2002eigenskin}
Kry, P.G., James, D.L., Pai, D.K.: Eigenskin: real time large deformation character skinning in hardware. In: Proceedings of the 2002 ACM SIGGRAPH/Eurographics symposium on Computer animation. pp. 153--159 (2002)

\bibitem{kynkaanniemi2022role}
Kynk{\"a}{\"a}nniemi, T., Karras, T., Aittala, M., Aila, T., Lehtinen, J.: The role of imagenet classes in fr$\backslash$'echet inception distance. arXiv preprint arXiv:2203.06026  (2022)

\bibitem{games}
Lefebvre, S., Hoppe, H.: Appearance-space texture synthesis. ACM Transactions on Graphics (TOG)  \textbf{25}(3),  541--548 (2006)

\bibitem{lefebvre2006appearance}
Lefebvre, S., Hoppe, H.: Appearance-space texture synthesis. ACM Transactions on Graphics (TOG)  \textbf{25}(3),  541--548 (2006)

\bibitem{syncTexture}
Liu, Y., Xie, M., Liu, H., Wong, T.T.: Text-guided texturing by synchronized multi-view diffusion. arXiv preprint arXiv:2311.12891  (2023)

\bibitem{liu2021swin}
Liu, Z., Lin, Y., Cao, Y., Hu, H., Wei, Y., Zhang, Z., Lin, S., Guo, B.: Swin transformer: Hierarchical vision transformer using shifted windows. In: Proceedings of the IEEE/CVF international conference on computer vision. pp. 10012--10022 (2021)

\bibitem{x-mesh}
Ma, Y., Zhang, X., Sun, X., Ji, J., Wang, H., Jiang, G., Zhuang, W., Ji, R.: X-mesh: Towards fast and accurate text-driven 3d stylization via dynamic textual guidance. In: Proceedings of the IEEE/CVF International Conference on Computer Vision. pp. 2749--2760 (2023)

\bibitem{Text2Mesh}
Michel, O., Bar-On, R., Liu, R., Benaim, S., Hanocka, R.: Text2mesh: Text-driven neural stylization for meshes. In: Proceedings of the IEEE/CVF Conference on Computer Vision and Pattern Recognition. pp. 13492--13502 (2022)

\bibitem{clip-mesh}
Mohammad~Khalid, N., Xie, T., Belilovsky, E., Popa, T.: Clip-mesh: Generating textured meshes from text using pretrained image-text models. In: SIGGRAPH Asia 2022 conference papers. pp.~1--8 (2022)

\bibitem{t2i-adapter}
Mou, C., Wang, X., Xie, L., Zhang, J., Qi, Z., Shan, Y., Qie, X.: T2i-adapter: Learning adapters to dig out more controllable ability for text-to-image diffusion models. arXiv preprint arXiv:2302.08453  (2023)

\bibitem{oechsle2019texture}
Oechsle, M., Mescheder, L., Niemeyer, M., Strauss, T., Geiger, A.: Texture fields: Learning texture representations in function space. In: Proceedings of the IEEE/CVF International Conference on Computer Vision. pp. 4531--4540 (2019)

\bibitem{pineda1988parallel}
Pineda, J.: A parallel algorithm for polygon rasterization. In: Proceedings of the 15th annual conference on Computer graphics and interactive techniques. pp. 17--20 (1988)

\bibitem{podell2023sdxl}
Podell, D., English, Z., Lacey, K., Blattmann, A., Dockhorn, T., M{\"u}ller, J., Penna, J., Rombach, R.: Sdxl: Improving latent diffusion models for high-resolution image synthesis. arXiv preprint arXiv:2307.01952  (2023)

\bibitem{dreamfusion}
Poole, B., Jain, A., Barron, J.T., Mildenhall, B.: Dreamfusion: Text-to-3d using 2d diffusion. arXiv preprint arXiv:2209.14988  (2022)

\bibitem{portenier2020gramgan}
Portenier, T., Arjomand~Bigdeli, S., Goksel, O.: Gramgan: Deep 3d texture synthesis from 2d exemplars. Advances in Neural Information Processing Systems  \textbf{33},  6994--7004 (2020)

\bibitem{clip}
Radford, A., Kim, J.W., Hallacy, C., Ramesh, A., Goh, G., Agarwal, S., Sastry, G., Askell, A., Mishkin, P., Clark, J., et~al.: Learning transferable visual models from natural language supervision. In: International conference on machine learning. pp. 8748--8763. PMLR (2021)

\bibitem{ramesh2022dalle2}
Ramesh, A., Dhariwal, P., Nichol, A., Chu, C., Chen, M.: Hierarchical text-conditional image generation with clip latents. arXiv preprint arXiv:2204.06125  \textbf{1}(2), ~3 (2022)

\bibitem{Texture}
Richardson, E., Metzer, G., Alaluf, Y., Giryes, R., Cohen-Or, D.: Texture: Text-guided texturing of 3d shapes. arXiv preprint arXiv:2302.01721  (2023)

\bibitem{sd}
Rombach, R., Blattmann, A., Lorenz, D., Esser, P., Ommer, B.: High-resolution image synthesis with latent diffusion models. In: Proceedings of the IEEE/CVF conference on computer vision and pattern recognition. pp. 10684--10695 (2022)

\bibitem{saharia2022photorealistic}
Saharia, C., Chan, W., Saxena, S., Li, L., Whang, J., Denton, E.L., Ghasemipour, K., Gontijo~Lopes, R., Karagol~Ayan, B., Salimans, T., et~al.: Photorealistic text-to-image diffusion models with deep language understanding. Advances in Neural Information Processing Systems  \textbf{35},  36479--36494 (2022)

\bibitem{clip-forge}
Sanghi, A., Chu, H., Lambourne, J.G., Wang, Y., Cheng, C.Y., Fumero, M., Malekshan, K.R.: Clip-forge: Towards zero-shot text-to-shape generation. In: Proceedings of the IEEE/CVF Conference on Computer Vision and Pattern Recognition. pp. 18603--18613 (2022)

\bibitem{savva2015semgeo}
Savva, M., Chang, A.X., Hanrahan, P.: {Semantically-Enriched 3D Models for Common-sense Knowledge}. CVPR 2015 Workshop on Functionality, Physics, Intentionality and Causality  (2015)

\bibitem{schuhmann2022laion}
Schuhmann, C., Beaumont, R., Vencu, R., Gordon, C., Wightman, R., Cherti, M., Coombes, T., Katta, A., Mullis, C., Wortsman, M., et~al.: Laion-5b: An open large-scale dataset for training next generation image-text models. Advances in Neural Information Processing Systems  \textbf{35},  25278--25294 (2022)

\bibitem{makeavideo}
Singer, U., Polyak, A., Hayes, T., Yin, X., An, J., Zhang, S., Hu, Q., Yang, H., Ashual, O., Gafni, O., et~al.: Make-a-video: Text-to-video generation without text-video data. arXiv preprint arXiv:2209.14792  (2022)

\bibitem{repaint3D}
Wang, T., Kanakis, M., Schindler, K., Van~Gool, L., Obukhov, A.: Breathing new life into 3d assets with generative repainting. In: Proceedings of the British Machine Vision Conference (BMVC). BMVA Press (2023)

\bibitem{wang2022diffusiondb}
Wang, Z.J., Montoya, E., Munechika, D., Yang, H., Hoover, B., Chau, D.H.: Diffusiondb: A large-scale prompt gallery dataset for text-to-image generative models. arXiv preprint arXiv:2210.14896  (2022)

\bibitem{wei2004tile}
Wei, L.Y.: Tile-based texture mapping on graphics hardware. In: Proceedings of the ACM SIGGRAPH/EUROGRAPHICS conference on Graphics hardware. pp. 55--63 (2004)

\bibitem{ying2001texture}
Ying, L., Hertzmann, A., Biermann, H., Zorin, D.: Texture and shape synthesis on surfaces. In: Rendering Techniques 2001: Proceedings of the Eurographics Workshop in London, United Kingdom, June 25--27, 2001 12. pp. 301--312. Springer (2001)

\bibitem{uv-diffusion}
Yu, X., Dai, P., Li, W., Ma, L., Liu, Z., Qi, X.: Texture generation on 3d meshes with point-uv diffusion. In: Proceedings of the IEEE/CVF International Conference on Computer Vision. pp. 4206--4216 (2023)

\bibitem{paint3d}
Zeng, X.: Paint3d: Paint anything 3d with lighting-less texture diffusion models. arXiv preprint arXiv:2312.13913  (2023)

\bibitem{zhang2023controlnet}
Zhang, L., Rao, A., Agrawala, M.: Adding conditional control to text-to-image diffusion models. In: Proceedings of the IEEE/CVF International Conference on Computer Vision. pp. 3836--3847 (2023)

\end{thebibliography}
\end{document}